\documentclass[10pt,twocolumn,letterpaper]{article}

\usepackage[pagenumbers]{cvpr} 

\usepackage{graphicx}
\usepackage{amsmath}
\usepackage{amssymb}
\usepackage{booktabs}
\usepackage{color}
\usepackage{multirow}
\usepackage[dvipsnames]{xcolor}
\usepackage{colortbl}
\usepackage{arydshln}
\usepackage{tabularx} 
\usepackage[accsupp]{axessibility}  

\usepackage[pagebackref,breaklinks,colorlinks]{hyperref}

\usepackage[capitalize]{cleveref}
\crefname{section}{Sec.}{Secs.}
\Crefname{section}{Section}{Sections}
\Crefname{table}{Table}{Tables}
\crefname{table}{Tab.}{Tabs.}

\def\cvprPaperID{1559} 
\def\confName{CVPR}
\def\confYear{2023}

\begin{document}

\title{StyleSync: High-Fidelity Generalized and Personalized Lip Sync \\ in Style-based Generator}

\author{Jiazhi Guan$^{1,2}$\thanks{Equal Contribution.} \quad Zhanwang Zhang$^{1}$\footnotemark[1] \quad Hang Zhou$^{1}$\thanks{Corresponding author.} \quad Tianshu Hu$^{1}$\footnotemark[2] \quad Kaisiyuan Wang$^{3}$   \\ Dongliang He$^{1}$ \quad Haocheng Feng$^{1}$ \quad Jingtuo Liu$^{1}$ \quad Errui Ding$^{1}$ \quad Ziwei Liu$^{4}$ \quad Jingdong Wang$^{1}$\\
$^1$Department of Computer Vision Technology (VIS), Baidu Inc. \quad $^2$Tsinghua University \\ $^3$The University of Sydney  \quad $^4$S-Lab, Nanyang Technological University\\
{\tt\small guanjz20@mails.tsinghua.edu.cn \{zhangzhanwang,zhouhang09,hutianshu01\}@baidu.com}
}

\twocolumn[{
\renewcommand\twocolumn[1][]{#1}%
\maketitle
\vspace{-30pt}
\begin{center}
 \centering
 \includegraphics[width=0.95\textwidth]{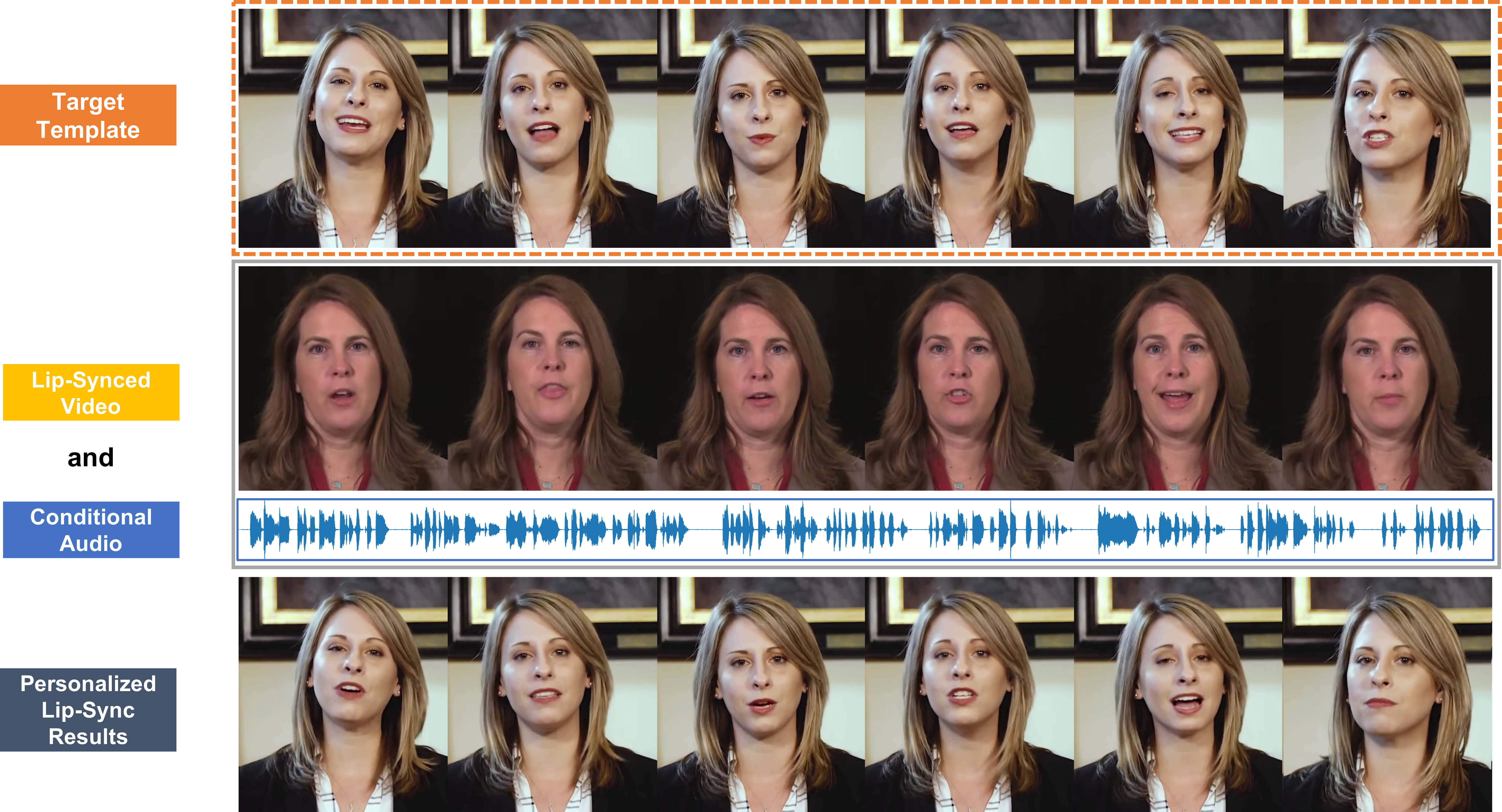}
\captionof{figure}{\textbf{Personalized lip-sync results generated by our StyleSync framework.} Our method not only supports high-fidelity modification to any target template video according to conditional audio but can further adapt to specific styles with personalized optimization. In this figure, our lip-sync results should have the same mouth shapes as the lip-synced video of the conditional audio.}
\label{fig:teaser}
\end{center}
}]

\maketitle
{
  \renewcommand{\thefootnote}%
    {\fnsymbol{footnote}}
  \footnotetext[1]{Equal contribution.}
  \footnotetext[2]{Corresponding authors.}
}

\begin{abstract}
   Despite recent advances in syncing lip movements with any audio waves, current methods still struggle to balance generation quality and the model's generalization ability. Previous studies either require long-term data for training or produce a similar movement pattern on all subjects with low quality. In this paper, we propose StyleSync, an effective framework that enables high-fidelity lip synchronization. We identify that a style-based generator would sufficiently enable such a charming property on both one-shot and few-shot scenarios. Specifically, we design a mask-guided spatial information encoding module that preserves the details of the given face. The mouth shapes are accurately modified by audio through modulated convolutions. Moreover, our design also enables personalized lip-sync by introducing style space and generator refinement on only limited frames. Thus the identity and talking style of a target person could be accurately preserved. Extensive experiments demonstrate the effectiveness of our method in producing high-fidelity results on a variety of scenes. Resources can be found at \url{https://hangz-nju-cuhk.github.io/projects/StyleSync}.
\end{abstract}


\section{Introduction}
\label{sec:intro}
The problem of generating lip-synced videos according to conditional audio is of great importance to the field of digital human creation, audio dubbing, film-making, and entertainment.
While  the rapid development of this area has been witnessed within recent years, most methods~\cite{zhou2019talking,chen2018lip,chen2019hierarchical,chen2020comprises,zhou2020makelttalk,zhou2021pose,ji2021audio,song2018talking,song2020everybody,wang2021audio2head,zhang2021facial,lu2021live,guo2021adnerf,lahiri2021lipsync3d,wu2021imitating,yi2020audio,ma2023styletalk} focus on generating a whole dynamic talking head. Results created under such settings can hardly be blended into an existing scene. 
Under real-world scenarios like audio dubbing,  one crucial need is to seamlessly alter the mouth or facial area while preserving other parts of the scene unchanged, making these methods non-feasible.

Previous methods take two different paths for achieving seamless mouth modification. A number of studies~\cite{song2020everybody,thies2020neural} pursue  realistic results on person-specific settings, which require long-term clips for target modeling. Moreover, they rely on prior 3D facial structural information. The uncertainty and errors accumulated in the 3D fitting procedure would greatly influence their performances. 
On the other hand, it is desired to build  models that break the data limitation on more generalized scenes. As a result, a few methods~\cite{prajwal2020lip,park2022synctalkface,sun2022masked} design person-agnostic models without relying on 3D or 2D structural priors. Nevertheless, such a setting is extremely challenging.

In order to produce high-fidelity lip-synced results on any-length videos, two essential challenges need to be addressed. \textbf{1)} How to efficiently design a powerful generative backbone network that supports both accurate audio information expression and seamless local alternation. Intuitively, the lip-sync quality naturally contradicts the preservation of the original target frame information~\cite{prajwal2020lip,zhou2021pose}. 
\textbf{2)} How to effectively leverage as much provided information as possible and involve the personalized properties to a generalized model. Though few-shot meta-learning has been proven effective in generating talking heads~\cite{zakharov2019few,burkov2020neural,chen2020talking,zhou2020makelttalk}, how to involve such ability into a lip-syncing pipeline has not been explored.

In this paper, we propose a highly concise and comprehensive framework named \textbf{StyleSync}, which produces high-fidelity lip-sync results on both generalized and personalized scenarios. The key is our \emph{simple but lip-sync-oriented modifications to style-based generator}. Though style-based generators~\cite{karras2019style, Karras2019stylegan2} have been leveraged in various talking head generation methods~\cite{burkov2020neural,zhou2021pose,yin2022styleheat}, their successes are only partially instructive. 
They aim at producing the whole head, which leads to unstable background and distortions which are non-acceptable in our scenarios. 

By revisiting the details of style-based generators, we identify a few simple but essential modifications that make our framework suitable for lip-syncing.
Different from the above methods, we adopt 
a masked mouth modeling protocol~\cite{prajwal2020lip,sun2022masked} and delicately design a \emph{Mask-based Spatial Information Encoding} strategy, where both the target and reference frames' information is encoded into a noise space~\cite{yang2021gan,xu2022styleswap} of the generator according to different masking schemes. While the information on audio dynamics and high-level reference frame is injected into the style-modulated convolution in a similar manner as~\cite{zhou2021pose,liang2022expressive}. In this way, our method can be benefited from the strong generative power of style-based generators and also keeps the advantage of easy implementation and fast training.

Moreover, our network modification enables personalized information preserving (e.g., speaking styles and details of the mouth and jaw). We take inspiration from the recent studies of inverting StyleGAN priors~\cite{abdal2019image2stylegan,abdal2020image2stylegan++,tov2021designing,roich2021pivotal} and propose a \emph{Personalized Optimization} scheme. As audio dubbing is normally performed on speaking videos, our model can make use of only a few seconds of the person's information and optimize additional person-specific 
parameters including the $W^+$ and the generator. Extensive experiments show that our framework clearly outperforms previous state of the arts on the one-shot setting by a large margin, and the target-specific optimization further enhances the fidelity of our results.

Our contributions can be summarized as follows: \textbf{1)} We present the \textbf{StyleSync} framework, which adopts simple but effective modifications including the \emph{Mask-based Spatial Information Encoding} to a style-based generator. \textbf{2)} We propose the \emph{Personalized Optimization} procedure which involves few-shot person-specific optimization into our framework. \textbf{3)}  Extensive experiments demonstrate that our framework can directly produce accurate and high-fidelity one-shot lip-sync results. Moreover, our proposed personalized optimization further improves the generation quality. Our method outperforms previous methods by a clear margin.
\section{Related Work}

\subsection{Audio-Driven Facial Animation}
The topic of audio-driven facial animation has long been a research interest in both the computer vision and graphics community. Studies have been carried out on both digital 3D human faces~\cite{karras2017audio,richard2021meshtalk,zhou2018visemenet,fan2022faceformer} and realistic human heads~\cite{fan2015photo,suwajanakorn2017synthesizing,jamaludin2019you}. We focus on human heads in the real world.

\noindent\textbf{Talking Head Generation.}
Most studies on lip-syncing target generating the whole head of a talking person~\cite{suwajanakorn2017synthesizing,jamaludin2019you,chen2018lip,zhou2019talking,zhou2021pose,chen2019hierarchical,chen2020comprises,yi2020audio,kaisiyuan2020mead,zhou2020makelttalk,guo2021adnerf,sun2021speech2talking,ji2022eamm,zhang2023dinet,wang2022one,shen2022dfrf,liang2022expressive,gururani2022spacex,liu2022semantic,ye2023geneface}. Specifically, a number of studies leverage  structural information such as 2D~\cite{chen2019hierarchical} landmarks, 3D landmarks~\cite{zhou2020makelttalk} and 3D meshes~\cite{chen2020comprises}. The uncertainty and inaccuracy of such representations would lead to error accumulation during the talking head generation procedure. Most person-specific methods~\cite{suwajanakorn2017synthesizing,lu2021live,ji2021audio} rely on them and produce results with a poor generalization or lip-sync quality.  Recently, NeRF is also used in person-specific modeling~\cite{guo2021adnerf,liu2022semantic,tang2022real,shen2022dfrf}, but they also perform poorly when limited data is provided.


\noindent\textbf{Lip-Syncing on Faces.}
The other type of study focuses on lip-syncing the mouth part while keeping other information untouched in videos~\cite{prajwal2020lip,park2022synctalkface,thies2020neural,song2018talking}. Our work lies within the same scope. While Thies et al.~\cite{thies2020neural} and Song et al.~\cite{song2020everybody}
produce realistic results, they rely on person-specific training on more than 2 minutes of data. 

Specifically, Wav2Lip~\cite{prajwal2020lip} generates person-agnostic lip-sync results that are highly accurate. However, they build the generative model on low-resolution images, leading to blurry results. We identify that it would be easier for the model to learn the mouth motions if less information about the image quality needs to be recovered. Moreover, their methods cannot capture the personalized patterns given the target template video. 

\subsection{Style-based Generator for Faces}
StyleGAN models~\cite{karras2019style,Karras2019stylegan2,Karras2021} have shown great success on image generation tasks, particularly on facial image generation and editing~\cite{abdal2019image2stylegan,abdal2020image2stylegan++,shen2020interfacegan,shen2021closedform,tov2021designing,richardson2021encoding}. 

\noindent\textbf{StyleGAN Inversion.}  StyleGAN inversion is the practice of recovering the latent space of a given image with a pre-trained StyleGAN. Abdal \textit{et al.}~\cite{abdal2019image2stylegan,abdal2020image2stylegan++} propose to invert images not only using the style space ${W}$, but also expand it to the $\mathcal{W^+}$, which better preserves details.
Recent studies propose to learn encoders~\cite{tov2021designing,alaluf2021restyle,richardson2021encoding} for better editability and faster optimization results. Furthermore, recent studies try to refine the generator's parameters through pivotal finetuning~\cite{roich2021pivotal}. In our work, we take inspiration of previous StyleGAN inversion studies and propose a personalized optimization procedure.  We encode a $\mathcal{W^+}$ space from reference frames following~\cite{tov2021designing} which enables personalized lip movements learning.

\label{sec:2.2}
\noindent\textbf{Applications.} StyleGAN architecture has also been leveraged in face restorations~\cite{wang2021gfpgan,yang2021gan} and face swapping~\cite{xu2022styleswap,xu2022region}. These tasks require preserving the original facial emotion and expressions. On the other hand, style-based generators have also been applied to create facial animations~\cite{burkov2020neural,zhou2021pose,liang2022expressive}, which are highly related to our task. However, they rely on the style vectors in the $W$ or $W^+$ space for controlling both the appearances and motion dynamics. One major drawback of this setting is that $W^+$ space cannot easily accounts for the spatial consistency of backgrounds, leading to non-realistic results or visible artifacts. 

\begin{figure*}[t]
    \centering
    \includegraphics[width=0.99\linewidth]{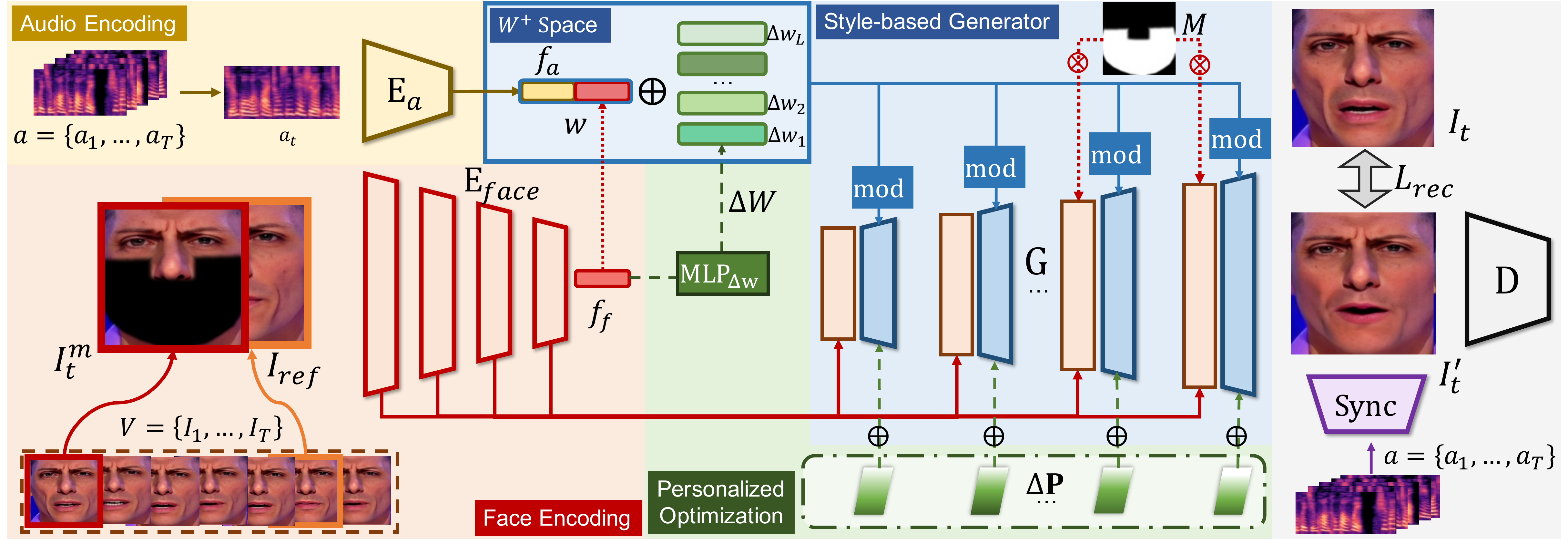}
    \caption{
   \textbf{Our StyleSync framework.} The building blocks in \textbf{Blue} indicate the style-based generator. The masked target frame $I^m_t$ is concatenated with a reference $I_{ref}$ and encoded to $f_{f}$ by $\text{E}_{face}$ (\textbf{Red}). The audio information is encoded to an audio feature $f_a$ (\textbf{Yellow}). The features are concatenated together to form the style ($W$) space.
   Specifically, we devise a Personalized Optimization procedure (\textbf{Green}) including the learning of the $\Delta \textbf{W}$ and the $\Delta \textbf{P}$. This part is not trained during the initial backbone learning.
}
\label{fig:pipeline}

\end{figure*}

\section{Methodology}

In this section, we introduce the details of our StyleSync framework as depicted in Fig.~\ref{fig:pipeline}. It is simply designed by leveraging several successes of previous style-based generators. We introduce our modifications to the style-based generator that make a successful backbone in Sec.~\ref{sec:3.1} and the training objectives of our generalized model in Sec.~\ref{sec:3.2}. Specifically, we illustrate our \emph{Personalized Optimization} procedure which further strengthens our results with person-specific optimization in Sec.~\ref{sec:3.3}.

\noindent\textbf{Task Formulation.} Our goal is to sync the lip motion of a target person with any provided audio and seamlessly blend it into the original target video. 
We formulate our training in a typical self-reconstruction manner. The training setting is similar to Wav2Lip~\cite{prajwal2020lip}. Given a target video $\textbf{V} = \{I_1, \dots, I_T\}$ with $T$ frames, we mask out the lower half of the face including the jaw and cheeks with a mask $M$. The goal is to recover the whole face with its corresponding audio $\textbf{a} = \{a_1, \dots, a_T\}$ ($\textbf{a}$ is processed to spectrograms). 
As no information about the mouth and jaw shape is provided on the masked target frame  $I^m_t = (\textbf{1} - M) * I_t$, we leverage a random reference frame $I_{ref} \in \textbf{V}$ during training to provide the desired context.

\subsection{Modifying Style-based Generator for Lip Sync}
\label{sec:3.1}

\noindent\textbf{Style-based Generator Overview.} We perform lip-sync-oriented modifications to build a successful backbone for lip sync with  the StyleGAN2~\cite{Karras2019stylegan2} architecture. The original StyleGAN2 takes a style vector $w$ as input and uses it to modulate  convolution operations in a total of $L$ generative layers during training. During inference, different $w$s can be used at different layers, formulating the $W^+$ space $\textbf{W} = \{w_1, \dots, w_L\}$. The basic elements of the original StyleGAN2, including the $W^+$ space and the StyleGAN generator $\text{G}$'s layers are depicted in \textbf{blue} on Fig.~\ref{fig:pipeline}. For simplification, certain detailed structures are omitted.


\noindent\textbf{Mask-based Spatial Information Encoding.} 
Our goal is to seamlessly blend a lip-synced mouth into the target frame with the assistance of a reference image. As discussed in Sec.~\ref{sec:2.2}, previous methods leveraging the style-based generator are not directly applicable to our scene, we seek a different way to encode the spatial information on faces. 

It is proven by recent studies on face restoration~\cite{yang2021gan} and swapping~\cite{xu2022styleswap} that content-guided feature maps can be attached to StyleGAN layers in a similar status as \emph{noise} without affecting the expressive power of the generative model.
These contextual-rich \emph{noises} $\textbf{N} = \{N_1, \dots, N_L\}$ could well preserve the original spatial structure and attributes of the encoded visual information. This property provides an option for leveraging the attributes information lying in the reference image. Thus following Wav2Lip~\cite{prajwal2020lip}, we concatenate the masked and reference input together as the visual input of the framework and encode the corresponding feature maps $\textbf{F}_f = \{F_1, \dots, F_L\}$. The simplest setting is to regard $\textbf{N} = \textbf{F}$.

However, the goal of generating lip motions  needs to create dynamics that are different from the original facial structure but keeps facial identity, which is substantially different from previous methods that aim to keep the structure fixed. 
The mouth shapes would be greatly influenced by the reference image $I_{ref}$ when adopting the same protocol. We identify that more than desired facial structure information is infused in the low-level information of $\textbf{F}$ (layers with higher resolutions). As a result, we intuitively mask out the low-level part of the information in the face features. 
Here we simply define 
\begin{align}
\label{eq:1}
F'_l = (\textbf{1} - M) * F_l, \text{for}~l > \lfloor \frac{L}{2} \rfloor.
\end{align}
The facial information is colored in \textbf{red} on Fig.~\ref{fig:pipeline}.
The $\textbf{1}$ is an all-one matrix that has the same size as $M$.

\noindent\textbf{Style Information Encoding.}
We follow previous studies~\cite{burkov2020neural,zhou2021pose,sun2021speech2talking} to encode both audio dynamics and facial information into the style space ($W$ space) of the style-based generator. The audio feature $f_a$ is encoded from the encoder $\text{E}_a$ and the face feature $f_f$ is the bottleneck of the face encoder $E_{face}$, $w = \text{concat}(f_a, f_f)$. As the information in the reference frames is already fused to the generator through spatial noise encoding, the fusing of the face feature is less necessary. Experiments show that the difference is subtle with or without $f_f$, but we still keep this design in accordance with common practice. 

\label{sec:3.1.3}
\noindent\textbf{Ingredients for Personalization.} Our above designs are not only simple but also possess external potentials of personalized lip-sync modeling. We leave the details to Sec.~\ref{sec:3.3} and briefly  discuss the additional ingredients that make our model comprehensive.

Specifically, our extension is inspired by the advances in StyleGAN inversion~\cite{abdal2019image2stylegan,abdal2020image2stylegan++,roich2021pivotal}. It  has pointed out that extending $w$ to the $W^+$ space which contains $\textbf{W} = \{w + \Delta w_1, \dots, w + \Delta w_L\}$ is essential to recovering specific images. Moreover, recent studies~\cite{roich2021pivotal,alaluf2022hyperstyle} also explore tuning the parameters of the generator to improve the network's fitting ability on a specific target. Thus our generator has the potential of learning limited parameter shifts $\Delta \textbf{P}$ in order to fit a specific person.

These personalized ingredients are depicted in \textbf{green} on Fig.~\ref{fig:pipeline} and are not optimized during the generalized training procedure. In order to keep the pipeline consistent across both settings, we set $\Delta \textbf{W} = \textbf{0}$ and $\Delta \textbf{P} = \textbf{0}$ in our generalized model.

\subsection{Backbone Training Objectives}
\label{sec:3.2}
During training, the whole backbone networks take the 6-channel concatenation of $[I^m_t, I_{ref}]$ and the audio clip  $a_t$ at the same time step as input. It predicts $I'_t = \text{G}(\textbf{N}, \textbf{W})$, which aims at recovering the unmasked target image $I_t$. 
In order to keep our design simple, the training objectives are mostly aligned with StyleGAN2~\cite{Karras2019stylegan2}
and Wav2Lip~\cite{prajwal2020lip}. 

\noindent\textbf{Reconstruction Loss.} The pixel-level reconstruction loss $\mathcal{L}_{rec}$ is fundamental for training our task. Here we leverage the commonly used $L_1$ loss and perceptual loss~\cite{wang2018high}.
\begin{align}
    \mathcal{L}_{rec} = \|I'_t - I_t\|_1 +     \sum_{m=1}^{N_{vgg}}\|\text{VGG}_m(I'_t) - \text{VGG}_{m}(I_t) \|_1,
\end{align}
where $\text{VGG}_m$ is the $m$th layer's output of a pre-trained VGG19 network.

\noindent\textbf{Adversarial Loss.}  We directly adopt the same discriminator $\text{D}$ from StyleGAN2~\cite{Karras2019stylegan2} for the adversarial training:
\begin{align}
    \mathcal{L}_{adv} =  \underset{\text{G}}{\text{min}}\underset{\text{D}}{\text{max}}(\mathbb{E}_{I_{t}}[\log \text{D}(I_{t})]  
     + \mathbb{E}_{I'_t}[\log (1 - \text{D}(I'_t))]).
\end{align}
Particularly, we initialize the discriminator's weights with the pre-trained version in StyleGAN2~\cite{Karras2019stylegan2}. This practice produces visibly sharper results.

\noindent\textbf{Lip-Sync Loss.} We additionally train a SyncNet~\cite{chung2016out} which consists of a visual encoder $\text{S}_v$ and an audio encoder $\text{S}_a$. It identifies whether visual and audio clips are timely aligned with contrastive loss. When training the generator, we predict 5 consecutive frames $I'_{t:t+4}$ within one batch at each inference step and supervise the training with SyncNet's assistance. The objective is:
\begin{align}
    \mathcal{L}_{sync} = - \frac{\text{S}_v(I'_{t:t+4})^\textbf{T} \cdot \text{S}_a(a_{t:t+4})}{\|\text{S}_v(I'_{t:t+4})\|_2\|\text{S}_a(a_{t:t+4})\|_2}
\end{align}

The overall loss functions across the generalized training can be written as:

\begin{align}
    \mathcal{L}_{g} =\mathcal{L}_{adv} + \lambda_r \mathcal{L}_{rec} + \lambda_s \mathcal{L}_{sync}.
\end{align}


\begin{figure*}[t]
\centering
\includegraphics[width=\linewidth]{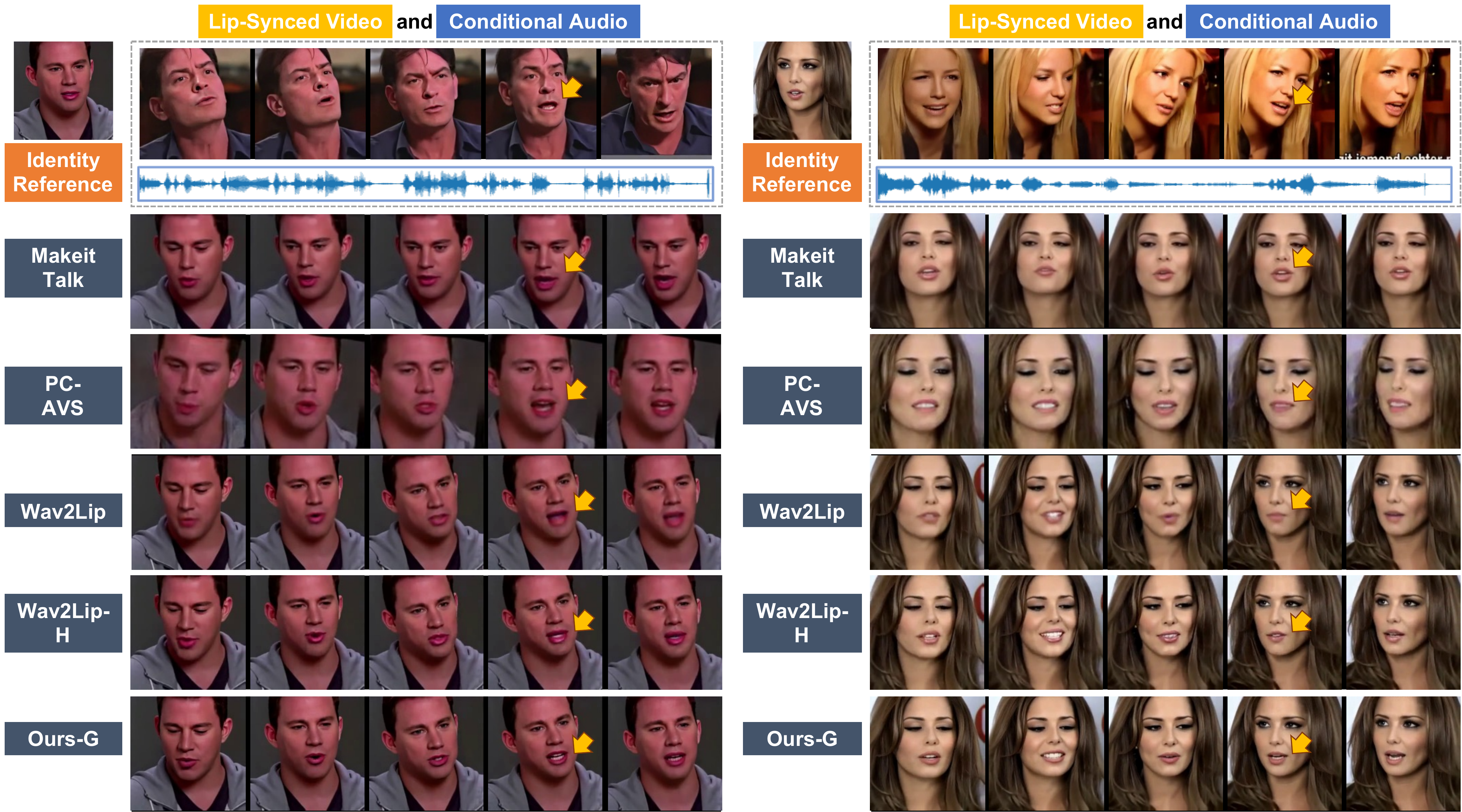}
\caption{\textbf{Qualitative Results}. The top row shows the lip-synced videos of the conditional driving audio.  MakeitTalk~\cite{zhou2020makelttalk} fails to generate accurate mouth shape and lacks head dynamics. PC-AVS~\cite{zhou2021pose} and Wav2Lip~\cite{prajwal2020lip} generate lip-synced results aligned well with audios. But they produce visibly blurry results. Particularly, our reproduced Wav2Lip-H generates high-quality results with good lip-sync. However, they still produce certain artifacts. While our generalized model directly produces high-fidelity results that have the same lip motion as the original synced video.}
\label{fig:exp_show_cases}
\end{figure*}


\subsection{Personalized Optimization}
\label{sec:3.3}
After the training of the backbone networks, our method can readily generate high-fidelity lip sync results for arbitrary subject. However,  as the model is trained in a generalized manner, the generated mouth motion patterns across different people are basically the same. It has been verified that different identities possess different talking styles~\cite{wu2021imitating,yi2020audio}. Here we pursue to capture such personalized property with the original template video. 
Notably, we focus on the few-shot setting that only less than one minute of the original video is given, which cannot be handled by previous person-specific models~\cite{thies2020neural,ji2021audio,song2020everybody,guo2021adnerf}.
%
Below, we illustrate how we successfully design our personalized optimization module based on the above discussions.

\noindent\textbf{Basic Learning Settings.}
The basic learning procedure of the personalized optimization is similar to backbone training. On a template video $\Bar{\textbf{V}} = \{\Bar{I}_1, \dots, \Bar{I}_T\}$ with audio $\Bar{\textbf{a}} = \{\Bar{a}_1 \dots, \Bar{a}_T\}$, we randomly select two frames as target $\Bar{I}_t$ and reference $\Bar{I}_{ref}$ for training as one sample. During training, the generalized loss $\mathcal{L}_{g}$ continuously supervises the whole procedure. 

As the encoder $\text{E}_{a}$ encodes person-agnostic speech content information, it is fixed during the personalized optimization step. While the face encoder $\text{E}_{face}$ encodes the attribute information of the target person, we do not optimize its parameters on a single person to avoid overfitting.

\noindent\textbf{Style Space Optimization.} According to our formulation, the $W^+$ space contains both the lip motion and the high-level facial style information. Thus optimizing it would ideally lead to person-specific lip motion pattern. E4E~\cite{tov2021designing} verifies that learning the small displacements around the original $W$ space would both extends the inversion quality of the image and enables strong editing ability. Thus we propose to learn a set of $\Delta \textbf{W} = MLP_{\Delta w}(f_f)$ in a similar way according to the encoded face feature $f_f$ with a few layers of MLPs $MLP_{\Delta w}$. 


\noindent\textbf{Generator Tuning.} The generator most accounts for the synthesizing and the blending quality. We allow the parameters $\textbf{P}$ of our generator $\text{G}$ to shift a little margin to $\Delta \textbf{P}$ with the personalized data we use. 

\noindent\textbf{Learning Objectives.} The final learning objectives of the personalized optimization procedure can be summarized as:
\begin{align}
    \mathcal{L}_{p} = \mathcal{L}_g + \lambda_p (\sum_{i,j} |\Delta  \textbf{W}^j_i|^2_2 + \sum_{m,n} |\Delta  \textbf{P}^m_n|^2_2),
\end{align}
where $\textbf{W}^j_i$ denote the $j$th parameter of the $i$th element in $\textbf{W}$.
This stands the same for $\textbf{P}^m_n$. 
We restrict the displacements in a limited step.

\section{Experiments}

\paragraph{Datasets.} We conduct experiments on two commonly used audio-visual datasets, LRW~\cite{chung2016lip} and VoxCeleb2~\cite{Chung18b}. 
We follow the datasets' original train/test split and train our model on a mixture of these two datasets.
\begin{itemize}
    \item \textbf{LRW~\cite{chung2016lip}} is an audio-visual dataset collected from BBC news for lip reading. It is one of the  earliest large-scale audio-visual datasets with high quality. It consists of 1,000 one-second utterances in 500 words.
    \item \textbf{VoxCeleb2}~\cite{Chung18b} VoxCeleb~\cite{Nagrani17} and VoxCeleb2 are large-scale audio-visual datasets collected for the speaker verification task. We re-download around one-fifth of VoxCeleb2 for training, which are of high quality. The official test set is processed with GPEN~\cite{yang2021gan} for evaluation. 
\end{itemize}

\noindent\textbf{Implementation Details.}
We process all videos at 25 fps and align all faces according to pre-detected landmarks at the eyes. All faces are cropped to the size of $256 \times 256$. A same U-shape mask is adopted as shown in Fig.~\ref{fig:pipeline} to erase the mouth, cheeks and jaws at the target frame. All audios are processed in the same way as Wav2Lip~\cite{prajwal2020lip}. 
In order to keep our design simple and avoid extra hyperparameter tuning, we adopt most settings from the previous studies~\cite{Karras2019stylegan2,yang2021gan,prajwal2020lip}. 
Specifically, the generator has a total of $L = 14$ style-convolution layers. $\lambda_r$ is empirically set to 10 and all other $\lambda$s are selected to 1. For personalized optimization, we train the person's data for 5 epochs. Long training would not lead to better results. Particularly, we find that training the personalized model together with the general training data leads to more stable results, the portion of personal data and general data is set as 1:1 in our experiments.

\paragraph{Comparing Methods.}
As our model is built based on the person-agnostic setting, we compare our StyleSync framework with three state-of-the-art methods including, MakeitTalk~\cite{zhou2020makelttalk}, PC-AVS~\cite{zhou2021pose} and \textbf{Wav2Lip}~\cite{prajwal2020lip}.  \textbf{MakeitTalk}~\cite{zhou2020makelttalk} synthesizes talking head videos with natural head pose. \textbf{PC-AVS}~\cite{zhou2021pose} a pose source video to achieve pose-controllable talking face generation. While
MakeitTalk and PC-AVS  produce a whole talking head, Wav2Lip which focuses on the mouth part is our main competing method. As Wav2Lip is originally trained on a low-resolution with small networks, we carefully reproduce it on the same data and even loss functions as our methods. We denote this model as \textbf{Wav2Lip-H} (High-Quality). For fair comparison, all basic comparisons are carried out on the generalized setting \textbf{Ours-G}.

We also perform personalized optimization on the test set, where each frame lasts less than 10 seconds for both datasets. The results are denoted as \textbf{Ours-P}. We equally perform the generator finetuning on the above methods, and all of them perform worse than their original models, thus the results are not listed below. Additionally, we compare our personalized setting with ADNeRF~\cite{guo2021adnerf}, a recent talking head advance using neural radiance fields~\cite{mildenhall2020nerf}.

\begin{table*}[t]
\centering
\resizebox{\linewidth}{!}{
\begin{tabular}{lcccccccccc}
\toprule
\multirow{2}{*}{Method} & \multicolumn{5}{c}{LRW}            & \multicolumn{5}{c}{VoxCeleb2}     \\ 
\cline{2-11} 
\noalign{\smallskip}
                        & SSIM $\uparrow$ & PSNR $\uparrow$ & LMD $\downarrow$ & $\text{Sync}_{conf}$ $\uparrow$ & $\mathcal{D}_{ID}$ $\uparrow$ & SSIM $\uparrow$ & PSNR $\uparrow$ & LMD $\downarrow$ & $\text{Sync}_{conf}$ $\uparrow$ & $\mathcal{D}_{ID}$ $\uparrow$ \\ 
\hline
\noalign{\smallskip}
MakeitTalk      &   0.69   &   29.83   &   2.75  &   3.88  &  0.79         &   0.63   &   28.38   &  6.94   &   2.15  & 0.71            \\
PC-AVS                  &   0.79   &   30.26   &   1.84  &   7.19   &  0.82       &   0.71   &   29.53   &  2.75   &   8.16   &  0.74         \\
Wav2Lip                 &   0.79   &   30.54   &   1.28  &   7.39   &  \textbf{0.90}        &   0.80   &   30.53   &  1.92   &   8.90    &   \textbf{0.90}     \\
Wav2Lip-H              &   {0.80}   &   31.38   &   1.20  &   7.19   &  0.88        &   \textbf{0.81}   &   30.53   &  {1.87}   & 8.35  & \textbf{0.90} \\
GT                      &   1.00   &   100.00   &  0.00   &  7.65  &   1.00       &   1.00   &   100.00   &  0.00   &  7.71  &    1.00 \\
\hline
\noalign{\smallskip}
Ours-G            &   \textbf{0.85}   &   \textbf{31.78}   &  \textbf{1.18}   &  7.25   &  0.89         &   0.79   &   \textbf{31.00}   &  \textbf{1.47}  &  8.25  &  \textbf{0.90}         \\
Ours-P            &   \textbf{0.88}   &   \textbf{32.66}  &  \textbf{0.86}  &  6.35   &  \textbf{0.93}       &   \textbf{0.82}   &   \textbf{31.54}   &  \textbf{1.15}   &  7.26  &   \textbf{0.93}        \\
\bottomrule
\end{tabular}
}
\caption{\textbf{Quantitative results on LRW and VoxCeleb.} For LMD the lower the better, and the higher the better for other metrics.}
\vspace{-0.1cm}
\label{tab:exp}
\end{table*}







\subsection{Quantitative Evaluation}  

As quantitative evaluations can only be performed on the self-construction setting, we avoid directly leveraging all information within the frame for reconstruction. We uniformly select a same random frame as the reference frame for PC-AVS, Wav2Lip, Wav2Lip-H and ours.

\noindent\textbf{Evaluation Metrics.} We follow previous studies~\cite{chen2018lip,chen2019hierarchical,zhou2019talking,zhou2021pose} to adopt the popularly used  SSIM~\cite{wang2004image}, PSNR metrics for evaluating the generation quality, and the landmark distances around the mouth (LMD) and SyncNet's~\cite{chung2016out} confidence score to show the lip-sync quality. Please be noted that the SyncNet metric is computed on the officially released version, which is different from our implementation in the loss functions. 

Particularly, as the details of the target person cannot be fully recovered, we leverage the ArcFace~\cite{deng2019accurate} network and compute the frame-by-frame feature cosine distances $\mathcal{D}_{ID}$ for evaluating whether the generated results preserve the identity well.

\noindent\textbf{Evaluation Results.}
The quantitative experiments are carried out on the test set of LRW~\cite{chung2016lip} and VoxCeleb2~\cite{Chung18b} datasets. Please refer to Table~\ref{tab:exp} for the results. 
It can be seen that on the LRW dataset, the generalized version of our StyleSync already outperforms previous methods by a large margin for generation quality. While on the VoxCeleb2 dataset, the gap is less obvious. The reason might be that most training data of our methods are basically frontal view faces selected from LRW, while VoxCeleb2 contains more complicated scenes. On the other hand, after our personalized optimization, the performance of our model advances again. This shows the generated results are clearly more similar to the targets at this stage, which can also be verified by looking into the identity distances. 

Meanwhile, our method also achieves comparable performance on the lip-sync metrics on both datasets.
Our LMD score is slightly better than the competing methods. As for the SyncNet score, we achieve comparable results on LRW that are closer to the ground truth's SyncNet score. 
We argue it is meaningless to refer to the SyncNet scores for verifying the lip-sync quality once the metric has outperformed the ground truth. 
The $Sync_{conf}$ only reflects how well an audio-visual pair fits the learned SyncNet model rather than the true perceptual quality. 
Thus though generated results might outperform ground truth on the metric, it does not mean better sync quality.

After the personalization, the lip-sync score degrades. We assume that the specificity of the speaking style reduces the mouth opening level. This is also shown in Fig.~\ref{fig:exp_ablations}. The woman tends not to open her mouth wide, leading to smaller mouth movements

\subsection{Qualitative Evaluation}
Subjective evaluation is crucial in identifying the ability of generative models, particularly on videos. We strongly recommend readers to watch our \textbf{supplementary video} at \url{https://hangz-nju-cuhk.github.io/projects/StyleSync}.

Two cross-driven examples and their comparisons with the SOTA methods are shown in Fig.~\ref{fig:exp_show_cases}. We use an conditional audio from an arbitrary person selected from the test set to drive the template.
It can be seen that MakeitTalk~\cite{zhou2020makelttalk} cannot produce accurate lip movements. Moreover, both PC-AVS and Wav2Lip produce visible artifacts or blurry results.
Our reproduced Wav2Lip-H adopts the same training setting as ours, thus generates plausible results for most cases. Nevertheless, our generalized model still produces the most accurate mouth shapes with the highest fidelity.
%

\begin{figure}[]
\centering
\includegraphics[width=\linewidth]{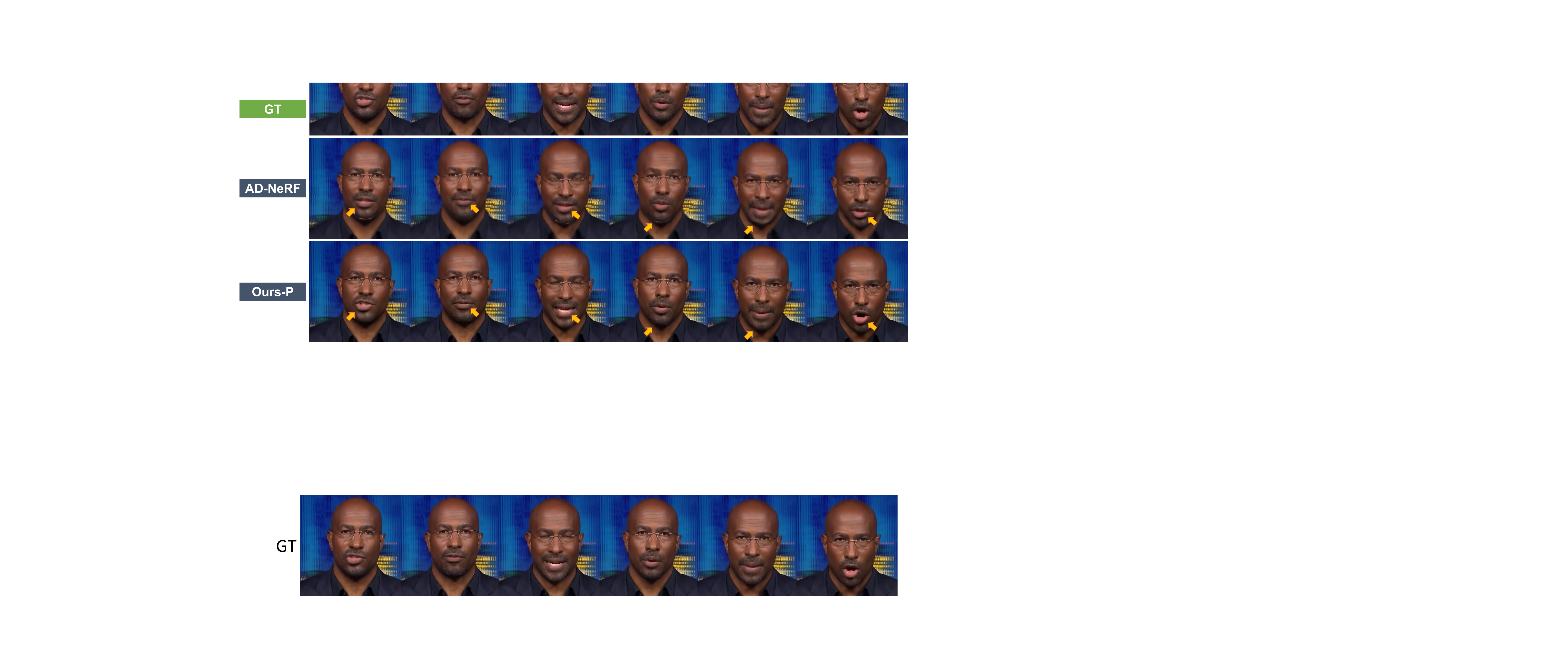}
\caption{{Self-driven results compared with personalized method.}}
\label{fig:cmp_adnerf}
\vspace{-8pt}
\end{figure}

Additionally, we show a case of personalized optimization on a 50-second video clip in Fig.~\ref{fig:teaser}. It can be seen that our method preserves the speaking style of the target person with accurate lip sync. 
Comparing our personalized results (Ours-P) with AD-NeRF \cite{guo2021adnerf} (Fig.~\ref{fig:cmp_adnerf}), it is evident that our method maintains more person-specific details with higher fidelity, even though we use significantly less personal data (10s for tuning v.s. 5min for training in \cite{guo2021adnerf}).

\noindent\textbf{User Study.}
We also invite 15 participants to conduct a user study for further subjective evaluation and the results are reported in Table~\ref{table:user_study}. Specifically, we randomly select 63 videos from LRW and VoxCeleb2 datasets as test sets and generate videos by using our generalized StyleSync and the comparison methods accordingly.
By adopting the commonly used Mean Opinion Scores (MOS) rating protocol, we request all the participants
to provide their ratings (from 1 to 5, the higher the better) on three aspects for each generated video: (1) Lip-Sync quality; (2) Generation quality; (3) Video realness.

As shown in Table~\ref{table:user_study}, MakeitTalk~\cite{zhou2020makelttalk} achieves the lowest scores in all the aspects due to lacking reasonable lip movements and head pose. PC-AVS~\cite{zhou2021pose}, Wav2Lip~\cite{prajwal2020lip} and Wav2Lip-H achieve relatively higher lip-sync score, but only Wav2Lip-H is able to synthesize videos with less blurry textures around the mouth.
Overall, our StyleSync outperforms its counterparts in all the three aspects by a large margin, indicating the effectiveness of our approach.

\begin{table}[] 
\centering
\resizebox{\linewidth}{!}{
\begin{tabular}{cccccc}
\toprule
MOS on $\setminus$ Approach & \rotatebox[origin=c]{60}{MakeitTalk} & \rotatebox[origin=c]{60}{PC-AVS} & \rotatebox[origin=c]{60}{Wav2Lip} & \rotatebox[origin=c]{60}{Wav2Lip-H} & \rotatebox[origin=c]{60}{\textbf{Ours-G}} \\
\midrule
Lip-Sync Quality  & 2.06 & 3.00 & 3.49 & 3.67 & \textbf{4.24}\\
Generation Quality & 2.63 & 2.16 & 1.87 & 3.42 & \textbf{4.52}\\
Video Realness     & 1.89 & 2.16 & 2.17 & 2.98 & \textbf{4.06}\\
\bottomrule
\end{tabular}
}
\caption{{User study measured by Mean Opinion Scores.} The scores are ranged from 1 (worst) to 5 (best).}
\label{table:user_study}
\vspace{-5pt}
\end{table}

\begin{table}[t]
\centering
\resizebox{\linewidth}{!}{
\begin{tabular}{lccccc}
\toprule
Method       & SSIM $\uparrow$ & PSNR $\uparrow$  & LMD $\downarrow$ & $\text{Sync}_{conf}$ $\uparrow$ & $\mathcal{L}_{ID}$ $\uparrow$ \\ \hline
w/o mask     & 0.81     &  31.20    &  1.80   &  7.89    & 0.90   \\
w/o sync     &   0.80   &  31.34    &  1.55   &  7.41    & 0.90   \\
Ours-G & 0.79   &   31.00   &  \textbf{1.47}   &  \textbf{8.25}  &  0.90    \\
\midrule
P w/o $\Delta \textbf{W}$ &  0.82    &  31.51    & 1.32    &  7.31    & 0.92   \\
Ours-P & 0.82   &   \textbf{31.54}   &  \textbf{1.15}   &  7.26  & \textbf{0.93}  \\
\bottomrule
\end{tabular}
}
\caption{Quantitative ablations results on VoxCeleb2.}
\vspace{-8pt}
\label{tab:ablation}
\end{table}


\subsection{Ablation Study}
To further demonstrate the contributions of our novel designs, we perform an ablation study on VoxCeleb2 dataset under both generalized setting and personalized optimization setting (denoted as ``Ours-G'' and ``Ours-P'', respectively). Concretely, we construct two variants for ``Ours-G'' (denoted as ``w/o mask'' and ``w/o sync'') by removing the masking operation in Eq.~\ref{eq:1} and the lip-sync loss during training, respectively.
While for ``Ours-P'', we form one variant (denoted as ``Ours-P w/o $\Delta \textbf{W}$'') by setting $\Delta \textbf{W} = \textbf{0}$. We have also experimented personalized optimization with fixed generated, however, this would lead to blurry results. This ablation is omitted here.

\begin{figure}[t]
\centering
\includegraphics[width=\linewidth]{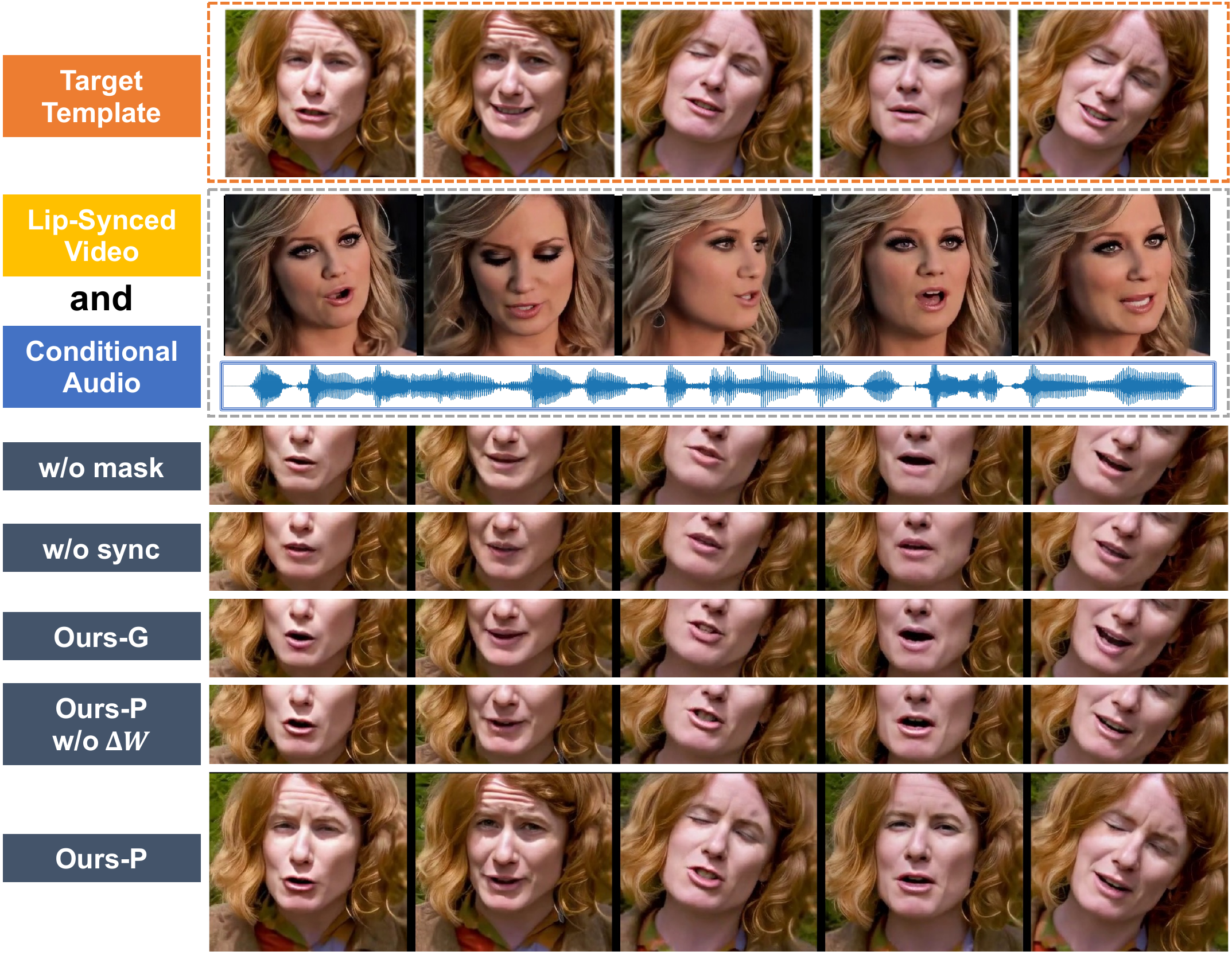}
\caption{{Ablation study with visual results.} Zoom in for details. }
\label{fig:exp_ablations}
\vspace{-8pt}
\end{figure}

The quantitative and qualitative results are shown in Table~\ref{tab:ablation} and Fig.~\ref{fig:exp_ablations}. As shown in Table~\ref{tab:ablation}, we observe that both ``w/o mask'' and ``w/o sync'' achieve comparable performance on the image quality metrics but leads to a performance drop on the lip-sync metrics.
``w/o mask'' achieves worse scores when compared with ``Ours-G'' in terms of LMD and Sync, which indicates that the proposed masking strategy can obviously alleviate the influence from the reference frames. While ``w/o sync'' reasonably suffers from lip-sync degradation and the results in Fig.~\ref{fig:exp_ablations} illustrate inconsistent lip movements when compared with the lip-synced video. The results demonstrate the effectiveness of the additional supervision from SyncNet.

It can be seen from Fig.~\ref{fig:exp_ablations} that with the personalized optimization, both the identity and the pattern of the mouth opening on the generated frames become more similar to the original template video. It is also clear that after the personalized training, the mouth opening is less obvious than in the generalized model. We analyze that this is part of the talking style of this target person. This also leads to poorer metric values on SyncNet.

In terms of the comparisons between our personalized model and ``Ours-P w/o $\Delta \textbf{W}$'', it achieves similar scores on the image quality as well as lip-sync quality. However, the identity score without $\Delta \textbf{W}$ is slightly degraded, which indicates that it is beneficial to involve $\Delta \textbf{W}$ into the training procedure for better preserving identity.

\section{Conclusion and Discussions}

\noindent\textbf{Conclusion.} In this paper, we propose \textbf{StyleSync}, which produces high-fidelity lip sync results for both the one-shot and the few-shot settings. We highlight the unique proprieties of our method: \textbf{1)}  Video results produced by our generalized model clearly outperform previous state-of-the-arts. \textbf{2)} Our model is built upon the success of recent style-based generators with simple modification. It is easy to implement and friendly to train. \textbf{3)} By involving StyleGAN-inspired personalized optimization procedure, our model can further be improved on a specific person given only a few clips.

\noindent\textbf{Limitations.} As our method blends a lip-synced face into an existing video with a fixed mask, the head pose and expressions of the target person cannot be changed. Additionally, under certain extreme cases where the target's jaw is extremely large, it might get out of our masked area. 

\noindent\textbf{Ethical Considerations.} Our method could be used to create non-existing talks and speeches, which might be maliciously used. We will issue our core models strictly to research institutions. 

\noindent\textbf{Acknowledgement.} This study is supported by the Ministry of Education, Singapore, under its MOE AcRF Tier 2 (MOE-T2EP20221-0012), NTU NAP, and under the RIE2020 Industry Alignment Fund – Industry Collaboration Projects (IAF-ICP) Funding Initiative, as well as cash and in-kind contribution from the industry partner(s).

{\small
\bibliographystyle{ieee_fullname}
\bibliography{egbib}
}

\end{document}